%% file: main_arxiv.tex
\pdfoutput=1

\documentclass{article}

\usepackage{arxiv}

\usepackage[utf8]{inputenc} %
\usepackage[T1]{fontenc}    %
\usepackage{hyperref}       %
\usepackage{url}            %
\usepackage{booktabs}       %
\usepackage{amsfonts}       %
\usepackage{nicefrac}       %
\usepackage{microtype}      %
\usepackage{lipsum}
\usepackage{graphicx}
\graphicspath{ {./images/} }

\usepackage{multirow}
\usepackage[ruled,vlined]{algorithm2e}
\usepackage{algorithmic}
\usepackage{algorithm2e}

\newcommand{\bda}{BD\textsuperscript{2}A\xspace}
\newcommand{\mil}{MIL\xspace} %
\newcommand{\tns}{TNSCUI\xspace}
\newcommand{\klo}{KlotskiNet\xspace}

\newtheorem{thm}{Theorem}
\newtheorem{pof}{Proof}

\newcommand*\numcircledmod[1]{\raisebox{.5pt}{\textcircled{\raisebox{-.9pt} {#1}}}}
\usepackage{subfigure}
\usepackage[capitalize]{cleveref}
\crefname{section}{Sec.}{Secs.}
\Crefname{section}{Section}{Sections}
\Crefname{table}{Table}{Tables}
\crefname{table}{Tab.}{Tabs.}
\Crefname{algorithm}{Alg.}{Algs.}

\newcommand{\wanglj}[1]{#1}
\newcommand{\shijie}[1]{#1}

\title{Intrinsic Bias Identification on Medical Image Datasets}

\author{
  Shijie Zhang \\
  Huawei Cloud Computing Technologies Company Limited\\
  \texttt{zhangshijie@pku.edu.cn} \\
   \And
  Lanjun Wang \\
  Tianjin University\\
  \texttt{wanglanjun@tju.edu.cn} \\
  \And
 Lian Ding \\
  Huawei Cloud Computing Technologies Company Limited\\
  \texttt{dinglian@huawei.com} \\
   \And
  An-an Liu \\
  Tianjin University\\
  \texttt{anan0422@gmail.com} \\
  \And
 Senhua Zhu \\
  Huawei Cloud Computing Technologies Company Limited\\
  \texttt{zhusenhua@huawei.com} \\
    \And
 Dandan Tu \\
  Huawei Cloud Computing Technologies Company Limited\\
  \texttt{tudandan@huawei.com} \\
}

\begin{document}
\maketitle
\begin{abstract}
Machine learning based medical image analysis highly depends on datasets. Biases in the dataset can be learned by the model and degrade the generalizability of the applications. 
There are studies on debiased models.
However, scientists and practitioners are difficult to identify implicit biases in the datasets,
which causes lack of reliable unbias test datasets to valid models.
To tackle this issue, we first define the data intrinsic bias attribute, and then propose a novel bias identification framework for medical image datasets. The framework contains two major components, KlotskiNet and Bias Discriminant Direction Analysis(\bda), where KlostkiNet is to build the mapping which makes backgrounds to distinguish positive and negative samples and \bda provides a theoretical solution on determining bias attributes. Experimental results on three datasets show the effectiveness of the bias attributes discovered by the framework. 
\end{abstract}

\input{Sections/01_Intro}

\input{Sections/02_relatedwork}
\input{Sections/03_prob}

\input{Sections/041_method_lanjun}

\input{Sections/052_exp_shijie}

\clearpage
\newpage  

\bibliographystyle{unsrt}  
\bibliography{reference}

\newpage  
\input{Sections/06_appendix}
\end{document}

%% file: Sections/01_Intro.tex
\section{Introduction} \label{sec:intro}

Machine learning has a significant role in several high impact applications in the medical imaging domains, such as computer-aided diagnosis, image segmentation, registration and fusion, image-guided therapy, image annotation, and image database retrieval. Example applications include the classification of interstitial lung diseases based on computed tomography (CT) images~\cite{anthimopoulos2016lung}, the classification of tuberculosis manifestation based on X-ray images~\cite{cao2016improving}, the classification of neural progenitor cells from somatic cell source~\cite{jiang2015convolutional}, etc.

Although medical imaging analysis has witnessed impressive progressing recent years thanks to the development of large-scale labeled datasets \cite{dataDCCL,datasets_xray,dataTNSCUI},  it may result in major discrimination if not dealt with proper care on biases within such data~\cite{fabbrizzi2021survey}. The bias in the data causes generalizability degradation of the developed models, which has been identified as one of major limitations in deep learning applications in healthcare~\cite{topol2019high}. For example, in the use case of apical lesion detection on panoramic radiographs, models trained only on Germany images showed a (mean±SD) F1-score of 54.1±0.8\% on Germany data but 32.7±0.8\% on Indian data ($p<0.001$/t-test)~\cite{krois2021generalizability}. 
\shijie{Recent studies have proposed several debiased models \cite{reddy2021benchmarking,barbu2019objectnet} but few practical applications in medical images, due to the unknown bias and lack of reliable test datasets.}

\begin{figure}
\centering  
\subfigure[Strip]{\includegraphics[width=3cm,height=2cm]{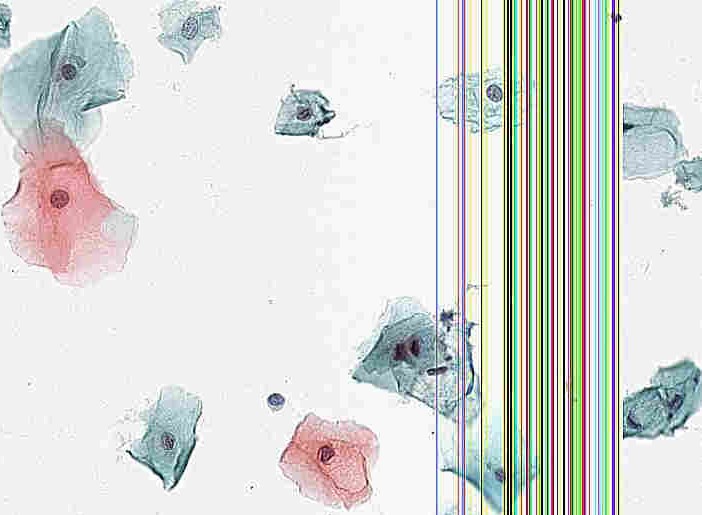}}
\hspace{0.5cm}
\subfigure[Inflammation]{\includegraphics[width=3cm,height=2cm]{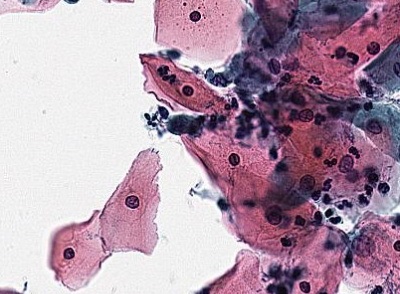}}%
\caption{Bias Examples}\label{fig:biasexp}
\end{figure}

In the community, scientists and practitioners have been aware of the biases existed in their datasets, but it is not a trivial problem to identify biases from medical image datasets.  There are two major challenges. 
Firstly, different from manually set biased attributes, like demographic properties(e.g., race, gender, etc.)~\cite{steed2021image}, or application specific properties \cite{bissoto2020debiasing,li2020shape}, biased attributes in medical image datasets have various types (i.e., too numerous to enumerate) and implicit representations (i.e., hard to model).  \cref{fig:biasexp} illustrates two cases. One of them is the strip might be caused by an accidental failure of the machine. Beside visible strips like in \cref{fig:biasexp}, in some cases, strips are invisible for human but can be recognized by neural networks. The other case is inflammation which appears in most of positive examples, but cannot be used as a feature in cancer diagnosis. 

Secondly, in medical images, bias features are coupled with true decision features (i.e. the features of lesions). Take inflammation in \cref{fig:biasexp} as an example, inflammation symptom does not only appear in the negative samples, but also in the positive regions of the positive samples.  This is different from the previous study in person re-identification where biases are only in the background \cite{tian2018eliminating}.  
As a result, we define the bias investigated in this study as data intrinsic bias because it is hidden in data. 

To tackle the above problems, we first define the data intrinsic bias attribute as the discriminant direction of the latent space where positive and negative samples are distinguishable without involving any information from lesions.  
Then, we propose a novel bias attribute identification framework for automatic bias discovery in medical datasets. 
The framework contains two major components, KlotskiNet and Bias Discriminant Direction Analysis (BD\textsuperscript{2}A). 
The objectives of proposed KlostkiNet is to build a mapping from the original space to the latent space. Specifically, different from ordinary classification tasks~\cite{campanella2019clinical,LI2019193,2019End}, KlostkiNet does not consider any tile with a lesion, but relies the tile with the maximum confidence to represent a whole sample in each training iteration. 
Moreover, BD\textsuperscript{2}A is designed to determine a data intrinsic bias attribute by solving  an optimal eigenvalue task, and a theoretical solution is provided.

To sum up, the major contributions of this study are as follows:
\begin{itemize}
    \item We provide a formal definition on the data intrinsic bias.  We call biases ``data intrinsic" is because they are hidden in the data.
    \item We propose a novel bias attribute identification framework, which contains two key components, KlotskiNet and BD\textsuperscript{2}A. KlotskiNet is designed to build up the bias embedding space and BD\textsuperscript{2}A is designed  to discover the discriminant direction of the space, which is the data intrinsic bias. 
    \item Experimental results on three datasets show the effectiveness of the bias attributes discovered by the framework. Compare with two baselines, the test set constructed by our framework can better \wanglj{expose the issues related to bias features in a model.}  
    \item We also provide a methodology to generate debiased test sets as a byproduct the framework for further algorithm evaluation.  
\end{itemize}

%% file: Sections/02_relatedwork.tex
\section{Related Work}\label{sec:rel}

Datasets are of central importance to computer vision and more broadly machine learning. 
As early as 2001, \cite{torralba2011unbiased} exposes the dataset bias problem in the general image recognition problem. 
More recently,  \cite{fabbrizzi2021survey} provides a survey of biases in visual datasets, where categorizes biases into selection bias, framing bias and label bias.  
However, in the medical image domain, it is not easy to classify biases into the above three types, because the biases come from a lot of different aspects, including machine, reagent, manipulation style of the doctor \cite{mohan2018mri}. 
For example, \cite{wachinger2021detect} studies the confounding bias which is from unwanted variability associated with scanners and sites when pooling data in neuroimaging datasets. 
Another example is a study of skin lesion detection using  ink markings/staining, and
patches applied to the patient skin as two of seven possible “culprit” artifacts for creating biases \cite{bissoto2020debiasing}.

Although researchers have been recognized biases in datasets, most studies are interested on manually set attributes which cause biases.  
\cite{bissoto2020debiasing} identifies seven artifacts (e.g.,  dark corners (vignetting), hair, ink markings/staining, etc.) which are possible sources of biases in the skin lesion detection, and finds out the spurious correlations exploited by biased networks.  
\cite{li2020shape} specifies that shape and texture are two prominent and complementary cues for recognizing objects and designs a debias method. 
Furthermore, \cite{steed2021image} focuses on human-like biases from social psychology (e.g. racial, gender, etc.) and observes that the model learns biases as the way people does. 

All the above studies expose bias issues in the dataset, but intrinsic biases are more important because they are unknown. %
A few  studies tackle the problem of bias attribute mining.  For example,
\cite{wachinger2021detect} applies Kolmogorov complexity to detect whether the confounded bias by providing the simplest factorization of the graphical model. However, this method can only be applied on tabular data.  

Besides, \cite{khosla2012undoing} and \cite{lopez2019dataset} address the problem of mining biases from image datasets.  These methods are based on the assumption that each dataset carries biases that make the provenance of its images easily distinguishable and affect the ability of the models to generalise well. Thus, they compare different datasets to identify biases.  
However, the assumption does not always exist in the medical image domain.  Firstly, it is possible that the bias distributions are same in multiple datasets (e.g., {two datasets have the same correlation between inflammation and cancer}), then these methods cannot be applied.  Secondly, due to multi-site, different geographic institutions have different positive phenotypes (i.e. {distributions} of true features  are not totally same), then the true features are identified as biases by the above methods. As a result, there is a need to mine intrinsic bias from medical image datasets.

In order to analyze, compare and evaluate the performance of algorithms, a debiased test dataset is necessary but not well addressed by previous studies.    
The most related work is ObjectNet\cite{barbu2019objectnet}, which is a test-only dataset with 50,000 images in 313 ImageNet classes, captured by crowdsourcing, following special guidelines to reduce biases, in particular, randomizing backgrounds, rotations, and image viewpoints. This dataset controls biases in the stage of data collection, but it costs a lot of human efforts.  
In this study, we provide a methodology to generate a debiased test dataset as a byproduct of data intrinsic bias mining for further algorithm evaluation.  %

%% file: Sections/03_prob.tex
\section{Problem Definition} \label{sec:prob}

Given a dataset $\mathcal{X}=\{(\mathbf{x}_1,y_1),\ldots,(\mathbf{x}_S,y_S)\}$, consists of $S$ training examples, $(\mathbf{x}_i,y_i)$, where $\mathbf{x}_i\in \mathbb{R}^M$ represents an $M$-dimension feature vector, and $y_i\in \{-1,1\}$ represents the label of the sample $i$. For a medical image dataset, the samples with the label $-1$ are called negative samples, denoted as $\mathcal{X}_N$, which do not contain any lesion. Meanwhile, the samples with the label $1$ are called positive samples, denoted as $\mathcal{X}_P$,  which contain lesions.

For a medical image $\mathbf{x}_i$, we can cut it into a set of $T$ tiles, denoted as  $\mathbf{R}_i=\{\mathbf{r}_{i,1},\ldots,\mathbf{r}_{i,T}\}$. Then, $\mathbf{R}_i$ can be divided into two subsets by whether containing foreground information: $\mathbf{R}_i^+$ and $\mathbf{R}_i^-$, where $\forall\ \mathbf{r}_{i,j}\in \mathbf{R}_i^+$, $\mathbf{r}_{i,j}$ contains foreground information of the example $\mathbf{x}_i$, and $\forall\ \mathbf{r}_{i,j}\in \mathbf{R}_i^-$, $\mathbf{r}_{i,j}$ only contains the background. It is noted that in a medical image, the lesion is the foreground. Then, for a positive sample $\mathbf{x}_i \in \mathcal{X}_p$, $\mathbf{R}_i=\mathbf{R}_i^+\cup\mathbf{R}_i^-$. Meanwhile, for a medical image $\mathbf{x}_i\in \mathcal{X}_N$, its corresponding foreground tile set is empty, i.e. $\mathbf{R}_i=\mathbf{R}_i^-$, because all tiles from a negative samples are without lesion. 

In this study, we focus on data intrinsic biases which make backgrounds to distinguish positive and negative samples. 
We set up a latent space with $K$-dimension and a map $g: \mathbb{R}^{M} \rightarrow \mathbb{R}^K$, i.e. $\mathbf{u}_{i}=g(\mathbf{x}_{i})$. 
In order to get rid of the true feature from foregrounds, the embedding aims to distinguish positive and negative samples by only using background tiles.  

We define a data intrinsic bias attribute as a direction $\phi$ in the latent space where the distributions of positive and negative samples are different in the one-dimensional subspace spanned by $\phi$.
Moreover, a data intrinsic bias feature is represented by a set of projected values by projecting samples on the subspace spanned by $\phi$, which is privileged with respect to one of labels.

Take a tabular data use case as an example to better understand data intrinsic bias attributes and bias features.  Given a job hunting scenario like~\cite{lahoti2019ifair}, where attributes are \textit{gender}, \textit{working experience}, and \textit{education experience}, and the \textit{salary} (high/low) is the label. Suppose in the dataset all workers with \textit{working experiences} over 10 years are labelled as high, then \textit{working experience} is a data intrinsic bias attribute, and \textit{working experience:$>10$} is the bias feature with respect to the high salary.

%% file: Sections/041_method_lanjun.tex
\begin{figure*}[t]
   \centering
   \includegraphics[width=0.7\linewidth]{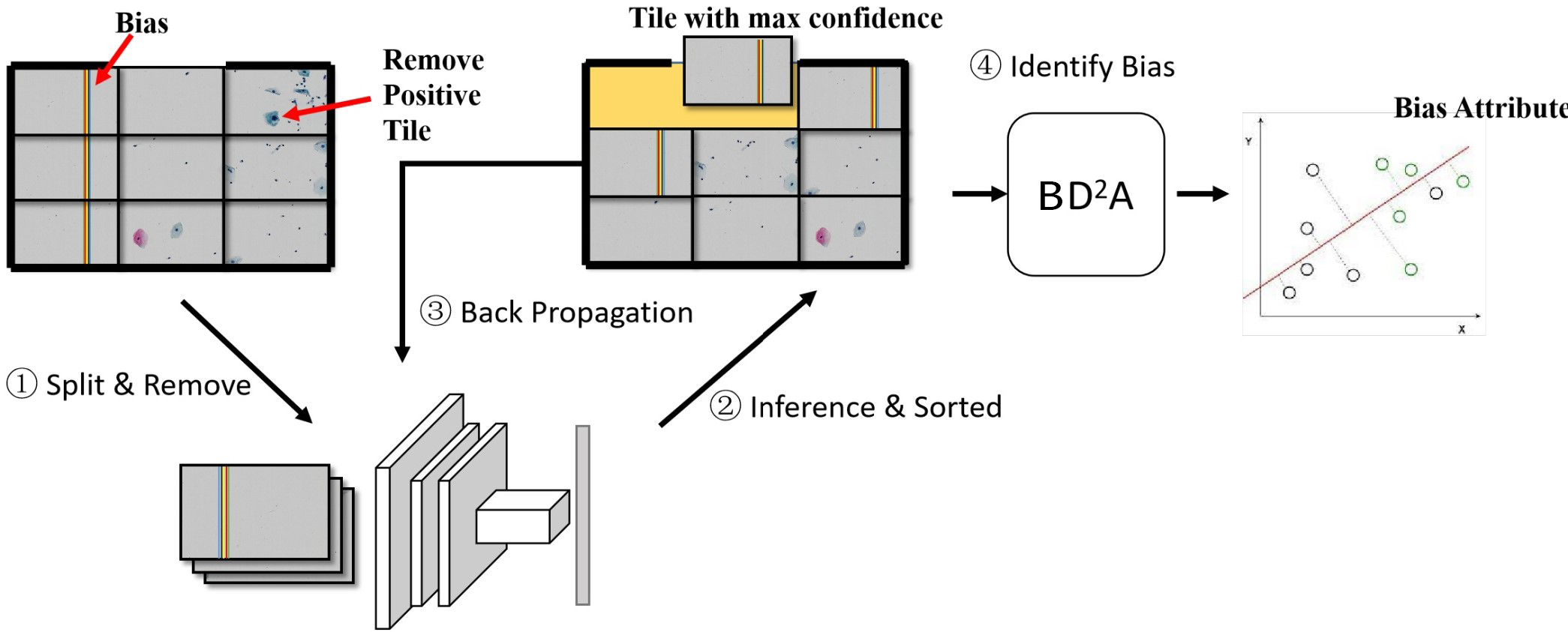}
  \caption{Overview of the intrinsic bias identification framework. \numcircledmod{1}Split the raw image evenly into $T$ tiles, and move out the tiles containing lesions.  \numcircledmod{2}Apply the model on each tile, output confidences and sort tiles by the max confidence. \numcircledmod{3}Pop out the key tile with the largest confidence and update the model by the output of the key tile. \numcircledmod{1}-\numcircledmod{3} illustrate the training process of KloskiNet.  After KloskiNet well trained, \numcircledmod{4}bias discriminant direction  analysis produces data intrinsic bias attributes (red line)}
   \label{fig:overview}
\end{figure*}
\section{Methodology} \label{sec:method}

\cref{fig:overview} provides an overview of our framework on mining data intrinsic bias attributes. In general, the framework contains two major components, KlotskiNet and Bias Discriminant Direction Analysis (\bda), where KlostkiNet is to build up the mapping $g$ which makes backgrounds to distinguish positive and negative samples and \bda~ is to obtain the discriminant direction of the embedding space. 

The KlotskiNet is a multi-instance learning algorithm to extract the embedding vectors from raw images (\cref{subsec:net}). In the training phase, KlotskiNet performs supervised training on the labeled data $\mathcal{X}$ by removing all positive tiles. In the inference phase, given an image $\mathbf{x}\in \mathcal{X}$ which is represented as its corresponding background tile set $\mathbf{R}^-$, KlotskiNet first makes predictions on each tile in $\mathbf{R}^-$, then select the one with the largest confidence, and finally computes the output of the last convolutional layer on the selected tile as the embedding of $\mathbf{x}$, denoted as $\mathbf{u}$. 

\bda~(\cref{subsec:bdv}) first formulates the definition of data intrinsic bias attribute identification task to a discriminant direction optimization problem with conjugated orthogonal constraints, and then provides a theoretical solution.

\subsection{KlotskiNet} \label{subsec:net}

We design a scheme called KlotskiNet to obtain a feature extractor to map an raw image to the latent space where positive and negative samples are distinguished by bias features.  \cref{algo:KlotskiNetTrain,algo:KlotskiNetInfer}  illustrate the training process and the feature extractor in the inference phase, respectively.

\cref{algo:KlotskiNetTrain} has two components, where Lines 1-8 is to obtain all negative tiles $\mathbf{R}^-_i$ by getting rid of the influence of foregrounds in the feature extractor, and Lines 9-17 is to training the network with the maximum confidence tile.  
Specifically, Line 11 is to obtain the confidence of each tile $\mathbf{r}_{i,j}$ in $\mathbf{R}^-_i$, where the tuple $(q^p_j, q^n_j)$ representing the normalized confidences of $\mathbf{r}_{i,j}$ on positive label and negative label, respectively.  
Lines 12-13 is to find out the index $j^*$ of the tile with the maximum confidence. We first get the indices of maximum positive and negative confidences, i.e. $j^*_p$ and $j^*_n$, respectively.  Then, to compare the maximum positive confidence and maximum negative confidence produces the index of the maximum confidence.   Finally,  Lines 14-15 illustrate to train the neural network with the cross-entropy loss calculated by the selected tile to distinguish positive and negative samples.

The reasons for selecting the tile with maximum confidence are twofold.  
Firstly, comparing the maximum confidence tile with a random one, the maximum one has much more significant features to make a higher confidence. As we have removed all positive tiles, those features are potential bias features.  
Secondly, we select the larger one between positive confidence and negative confidence.  Then, among the whole training set, the selected tiles contain both predicted positives and predicted negatives. Therefore, the model considers both positive and negative bias features. 

\begin{algorithm}
  \caption{KlotskiNet: Training Phase}
  \label{algo:KlotskiNetTrain}
  \begin{algorithmic}[1]
  \REQUIRE Raw image set $\mathcal{X}=\{(\mathbf{x}_1,y_1),\ldots,(\mathbf{x}_S,y_S)\}$
  \ENSURE KlotskiNet $f(\mathbf{r};\theta)$
  \\\COMMENT{\# Remove positive tiles}
  \FOR{$i \in \{1,\ldots,S\}$}
    \STATE $\mathbf{R}_i \gets Split(\mathbf{x}_i)$
    \IF{$y_i == 1$}
        \STATE $\mathbf{R}_i^-\gets$ $\mathbf{R}_i\setminus\{\mathbf{r}_{i,j}|lesion\ in\ \mathbf{r}_{i,j}\}$
    \ELSE
        \STATE $\mathbf{R}_i^-\gets\mathbf{R}_i$
    \ENDIF
  \ENDFOR
  \\\COMMENT{\# Training}
  \WHILE{conditions to continue training}
    \FOR{$i \in \{1,\ldots,S\}$}
        \STATE $\{(q^p_j,q^n_j)\}\gets \{f(\mathbf{r}_{i,j};\theta)|\mathbf{r}_{i,j} \in \mathbf{R}_i^-\}$
        \STATE $j^*_p \gets argmax \{q^p_{j}\}$,  $j^*_n \gets argmax \{q^n_{j}\}$
        \STATE $j^* \gets j^*_p$ if $q^p_{j^*_p} > q^n_{j^*_n} $ otherwise $j^*_n$
        \STATE Calculate cross-entropy loss with $y_i$ and $(q^p_{j^*}, q^n_{j^*})$
        \STATE Back propagation to update $\theta$
    \ENDFOR
  \ENDWHILE
  \end{algorithmic}
\end{algorithm}

\cref{algo:KlotskiNetInfer} is the process of extracting the embedding feature to distinguish positive and negative samples by KlotskiNet. Similar as the training process in \cref{algo:KlotskiNetTrain}, Lines 1-6 remove all positive tiles.  
It is noted that when we continue on \bda~ as \cref{subsec:bdv}, the training set are used to collect the samples in the embedding space, where labels are necessary because we want to identify bias attributes for positive and negative separately. 
However, when the purpose is to check whether a given image containing a bias feature,  it is necessary to project the sample to the latent space by \cref{algo:KlotskiNetInfer}, but the label is not necessary.   
Then, Lines 7-9 obtain the tile with the maximum confidence.   Finally, the embedding vector is generated by mapping the selected tile to the feature map of the last layer before the output layer.   

\begin{algorithm}
  \caption{Feature Extractor by KlotskiNet}
  \label{algo:KlotskiNetInfer}
  \begin{algorithmic}[1]
  \REQUIRE A raw image with its label $(\mathbf{x}, y)$, KlotskiNet $f(\mathbf{r})$ and the mapping representing its last layer before output layer $f'(\mathbf{r})$ 
  \ENSURE Embedding $\mathbf{u}$ of the input $\mathbf{x}$
  \\\COMMENT{\# Remove positive tiles}
  \STATE $\mathbf{R} \gets Split(\mathbf{x})$
    \IF{$y == -1$ or without label}
        \STATE $\mathbf{R}^-\gets\mathbf{R}$
    \ELSE
         \STATE $\mathbf{R}^-\gets$ $\mathbf{R}\setminus\{\mathbf{r}_{j}|lesion\ in\ \mathbf{r}_{j}\}$
    \ENDIF
    \\\COMMENT{\# Obtain the embedding}
    \STATE $\{(q^p_j,q^n_j)\}\gets \{f(\mathbf{r}_{j};\theta)|\mathbf{r}_{j} \in \mathbf{R}^-\}$
    \STATE $j^*_p \gets argmax \{q^p_{j}\}$,  $j^*_n \gets argmax \{q^n_{j}\}$
    \STATE $j^* \gets j^*_p$ if $q^p_{j^*_p} > q^n_{j^*_n} $ otherwise $j^*_n$
    \STATE $\mathbf{u} \gets f'(\mathbf{r}_{j^*})$  
  \end{algorithmic}
\end{algorithm}

\wanglj{To sum up, the key advantage of KloskitNet is to learn an image with erased positive tiles but without to involve any new bias related to 1) whether there is an erased region; 2) location or shape of the erased region. This is because KlotskiNet breaks the image into tiles, and leverages multi-instance learning to select the representative tile as the model input. The model  is only trained by the selected tile (i.e. with the max confidence) at a time, but does not aware which region is.  That is to say, the model makes use of the whole image while avoiding to contain any information from the erased region.}

\subsection{\bda} \label{subsec:bdv}

As defined in \cref{sec:prob}, a data intrinsic bias attribute as a direction $\phi$ in the latent space obtained by \cref{algo:KlotskiNetTrain} where the distributions of positive and negative samples are different in the one-dimensional subspace spanned by $\phi$. Since it is hard to formulate the distribution differences by empirical data, we leverage sample moments to represent the characteristics of the distribution.  In this study, we address on the second moment following linear discriminant analysis(LDA)~\cite{martinez2001pca}.  However, we cannot apply LDA directly because LDA provides an overall space discriminant direction but cannot generate the discriminant direction with respect to a specific label. 

Suppose $\mathcal{U}$ is produced by applying \cref{algo:KlotskiNetInfer} on the training set $\mathcal{X}$,  based on the above intuition that the direction is related to the label, we divide $\mathcal{U}$ into two subsets based on the label, which are $\mathcal{U}_p$ and $\mathcal{U}_n$. Suppose $\bar{\mathbf{u}}_p$, and $\bar{\mathbf{u}}_n$ are the mean vectors of the positive and negative embeddings, respectively.

In the following discussion, we use the way to find the positive bias attributes to illustrate the method.  It is noted that we can obtain the negative bias attributes in the same way by only switching the footmarks $p$ and $n$.

The positive label scatter matrix $S_p$ and the scatter of negatives centered by the mean of positive samples, which represents how negative samples are far away from positives, i.e., $S_{pn}$ are defined as:
\begin{eqnarray}
S_{p}&=&\mathbb{E}\{(\mathbf{u}_p-\bar{\mathbf{u}}_p)^T(\mathbf{u}_p-\bar{\mathbf{u}}_p)\} \\
S_{pn}&=&\mathbb{E}\{(\mathbf{u}_n-\bar{\mathbf{u}}_p)^T(\mathbf{u}_n-\bar{\mathbf{u}}_p)\}
\end{eqnarray}
The criterion function can be defined as follows:
\begin{equation}
    L(\phi) = \frac{\phi^T S_{pn} \phi}{\phi^T S_p \phi} \label{con:object_func}
\end{equation}
where $\phi$ is an arbitrary vector in $\mathbb{R}^K$. 

The vector $\phi_1$ corresponding to maximum of $L(\phi)$ is the optimal discriminant direction.  This vector $\phi_1$ is the eigenvector corresponding to maximum eigenvalue of the following eigenequation:
\begin{equation}\label{con:r_vector_eigen}
    S_{pn}{\phi_1} = \lambda S_p {\phi_1}
\end{equation}

As introduced in \cref{sec:intro}, biases have various types, thus, the optimal discriminant direction is not enough. In the following, we figure out a way to compute the $k$-th optimal direction.  Suppose $k$ directions $\phi_1,\phi_2,\ldots,\phi_k$ $(1 \leq k \leq K)$ are obtained, then we can make the following transformation from $\mathbb{R}^K$ to $\mathbb{R}^k$ to obtain the feature corresponding to each direction:
\begin{equation}\label{con:transform}
    \mathbf{v} = \Phi_k^T
    \mathbf{u}
\end{equation}
where $\Phi_k = [\phi_1^T,\phi_2^T,\ldots,\phi_k^T]$.

In order to obtain uncorrelated features, following \cite{UDVpaper}, a series of conjugated orthogonality constraints to \cref{con:r_vector_eigen}, which are $\forall\ i \leq k$, then:
\begin{equation}\label{con:constrains}
    \phi_{k+1}^T S_p \phi_i = 0 
\end{equation}
That is to say, suppose a positive sample embedding $\mathbf{u}_p$ is transformed to $\mathbf{v}_p$ with \cref{con:transform}, where $\mathbf{v}_p=[v_{p,1},\dots,v_{p,i}, \dots,v_{p,j}, \dots, v_{p,k}]^T$, then for any two features $v_{p,i}$ and $v_{p,j}$, they are statistically uncorrelated based on \cref{con:constrains}.

Finally, we can compute the $\phi_{k+1}$ according to the follow theorem:
\begin{thm}
The uncorrelated discriminant attribute $\phi_{k+1}$ is the eigenvector correspoding to maximum eigenvalue of the following eigenequation:
\begin{equation}
    [
        I - 
        S_p \Phi_k^T 
       \Phi_k
    ]
    S_{pn} \phi_{k+1} 
    = \lambda S_p \phi_{k+1} \label{con:sol}
\end{equation}
where $I$ is identity matrix. 
\end{thm}

\begin{pof}
Let $\phi_{k+1}$ satisfy the following constrain related to its normalization form as a complementary of \cref{con:constrains}:
\begin{equation}\label{con:const_normal}
    \phi_{k+1}^T S_p \phi_{k+1} = 1 
\end{equation}
Transform \cref{con:object_func} by considering all constrains from \cref{con:constrains} and \cref{con:const_normal} as 
\begin{eqnarray}\label{con:lagrange}
L(\phi_{k+1}) &=& \phi_{k+1}^T S_{pn} \phi_{k+1} - \lambda (\phi_{k+1}^T S_p \phi_{k+1}-1)\nonumber \\ 
                &-& \sum_{i=1}^k \mu_i \phi_{k+1}^T S_p \phi_{i}
\end{eqnarray}
where $\lambda$ and $\mu_i$ ($i=1,\ldots,k$) are Lagrange multipliers.  The optimal point is performed by setting the partial
derivative of $L(\phi_{k+1})$ with respect to $\phi_{k+1}$ as zero, which is:
\begin{equation}\label{con:deri}
    2S_{pn}\phi_{k+1} - 2\lambda S_p \phi_{k+1} - \sum_{i=1}^k \mu_i S_p \phi_{i} = 0
\end{equation}
Multiply \cref{con:deri} by $\phi_{i}^T$ ($i=1,\ldots,k$), then we can obtain
\begin{equation}\label{con:mu}
    \mu_i = 2\phi_i^T S_{pn} \phi_{k+1}
\end{equation}
Denote $M = [\mu_1,\ldots,\mu_k]= {\Phi}_k S_{pn} \phi_{k+1}$.
Put \cref{con:mu} into \cref{con:deri}, then we have
\begin{eqnarray}
    S_{pn}\phi_{k+1} - \lambda S_p \phi_{k+1} - S_p\Phi_k^T M = 0 \\
     S_{pn}\phi_{k+1}  - S_p\Phi_k^T {\Phi}_k S_{pn} \phi_{k+1} = \lambda S_p\phi_{k+1}
\end{eqnarray}
Therefore, \cref{con:sol} is obtained.

\end{pof}

%% file: Sections/052_exp_shijie.tex
\section{Experiments} \label{sec:exp}
To validate our method, we first introduce our experiment settings in \cref{subsec:setting}. Then, \cref{subsec:ovelperf} shows the overall performance of bias attributes identified by the framework.  
Finally, we evaluate the performance of individual bias attributes quantitatively (\cref{subsec:ablation}) and qualitatively (\cref{subsec:casestudy}). 
\wanglj{In addition, an extra dataset, more case studies as well as the parameter analysis} will be shown in Appendix. %
\subsection{Experiment Settings}\label{subsec:setting}
\subsubsection{Datasets} \label{subsec:data}

We conduct our experiments on two popular medical image datasets: Deep Cervical Cytological Lesions(DCCL)~\cite{dataDCCL} and Thyroid Nodule Segmentation and Classification in Ultrasound Images 2020(TNSCUI)~\cite{dataTNSCUI}.  
\shijie{To verify the potential usage of our framework on the general image domain, we also test on CI-MNIST~\cite{reddy2021benchmarking}.  More detailed information on CI-MNIST are in Appendix. }

\begin{table}[]
\centering
\caption{Datasets Overview}\label{tab:datasets}
\begin{tabular}{@{}lrrrrr@{}}
\toprule
           & Train\# & Val\#   & Test\#  & Total\# & Accuracy \\ \midrule
DCCL       & 5,919  & 1,905  & 2,979  & 10,803 & 0.869           \\
TNSCUI & 2,186  & 729   & 729   & 3,644  & 0.750           \\
CI-MNIST  & 50,000 & 10,000 & 10,000 & 70,000 & 0.897           \\ \bottomrule
\end{tabular}
\end{table}

Specifically, it is worthy noted that the last row ``Accuracy'' in \cref{tab:datasets}, which represents the accuracy of the test set on \klo.  As illustrated in \cref{algo:KlotskiNetTrain}, there is no positive tile involving to train \klo.  Suppose there is no bias attribute in data, the network should work as a random generator (i.e., accuracy is around 0.5) because all input is noise which does not contain any information of the lesion/foreground. However, in all of our three datasets, the accuracy of \klo is above 0.5, which indicates the existence of biases.

\subsubsection{Baseline Methods} \label{subsec:baseline}

\wanglj{\textit{\textbf{Color:}} To compare with our method, we first set up a baseline representing manually bias identification method. In previous studies \cite{2015Structure,2019Fast}, researchers have found that normalizing unwanted color variations aids model performance in computational pathology, thus we use average  RGB values of each image as the baseline attributes to compare with our framework.
}\wanglj{
\textit{\textbf{ImageNet-C(IC):}} \cite{imagenetc} is a common robustness evaluation dataset, which has been used for debiased method evaluation in previous studies\cite{li2020shape,useimegenetc1}.
In this paper, we apply the generation method from \cite{imagenetc} on medical image datasets to compare with our framework. Three possible corruptions in medical images are selected for experiments: brightness(B), contrast(C), and jpeg compression(J). 
}
\subsubsection{Evaluation Metric} \label{subsec:metric}
Following \cite{reddy2021benchmarking,barbu2019objectnet}, we use the drop of performance to evaluate the effectiveness of the framework.  

In the evaluation, we first identify biased attributes by our method and baselines. Then, we \textit{project} the test set $\mathcal{X}_{test}$ on an attribute, \textit{select} samples with an absolute projected value in the top-$\theta$ percentile as the sample with bias feature, and \textit{put} them into the biased set, where $\theta$ is the parameter controlling the projected value to be recognized as the bias feature. With this approach, we can separate biased set with respect to the attribute from the rest, which will be used in \cref{subsec:ablation,subsec:casestudy}.  When to evaluate the overall performance(\cref{subsec:ovelperf}), the above \textit{project, select and put} operations iterate with top-$k$ optimal directions, where $k$ is a parameter to be discussed in Appendix too.  Finally, we obtain the biased set $\mathcal{X}_{biased}$. %

\wanglj{In order to evaluate the performance drop on the real prediction model, we have implemented a multiple instance model(\mil) of decision support systems as \cite{campanella2019clinical}.}
The performance drop is defined by the performance metric of MIL model on $\mathcal{X}_{test}\setminus\mathcal{X}_{biased}$ minus that on $\mathcal{X}_{biased}$. The larger performance drop, the more biased information leveraged by the model.  

The performance metrics we will use include accuracy, precision, recall and area under the ROC curve(ROC-AUC). %
In particular, as a positive bias attribute affects negative samples, when checking the performance of an individual positive attribute, the recall and precision are calculated with respect to negative samples, and vice versa.

\begin{table*}[!htb]
\centering
\caption{Overall Performance
}%
   \label{tab:overall} 
\centering
\begin{tabular}{@{}llllllll@{}}
\toprule
Datasets & Method    & \begin{tabular}[c]{@{}l@{}}Drop of\\      Accuracy\end{tabular} & \begin{tabular}[c]{@{}l@{}}Drop of\\      Positive\\      Precision\end{tabular} & \begin{tabular}[c]{@{}l@{}}Drop of\\      Positive\\      Recall\end{tabular} & \begin{tabular}[c]{@{}l@{}}Drop of\\      Negative\\      Precision\end{tabular} & \begin{tabular}[c]{@{}l@{}}Drop of\\      Negative\\      Recall\end{tabular} & \begin{tabular}[c]{@{}l@{}}Drop of\\      ROC-AUC\end{tabular} \\ \midrule
         & IC-B\cite{imagenetc}      & 0.124                                                           & -0.026                                                                           & 0.213                                                                         & -0.042                                                                           & 0.129                                                                         & 0.043                                                          \\
         & IC-BJ\cite{imagenetc}     & 0.174                                                           & 0.201                                                                            & 0.011                                                                         & 0.447                                                                            & 0.105                                                                         & 0.205                                                          \\
DCCL     & IC-BJC\cite{imagenetc}    & 0.198                                                           & 0.123                                                                            & 0.209                                                                         & 0.167                                                                            & 0.176                                                                         & 0.220                                                          \\
         & Color\cite{2015Structure,2019Fast}   & -0.050                                                          & -0.011                                                                           & -0.078                                                                        & -0.087                                                                           & -0.010                                                                        & -0.039                                                         \\
         & Our & \textbf{0.561}                                                  & \textbf{0.454}                                                                   & \textbf{0.584}                                                                & \textbf{0.573}                                                                   & \textbf{0.480}                                                                & \textbf{0.624}                                                 \\ \midrule
         & IC-B\cite{imagenetc}      & -0.023                                                          & 0.021                                                                            & -0.150                                                                        & 0.149                                                                            & -0.051                                                                        & -0.018                                                         \\
         & IC-BJ\cite{imagenetc}     & 0.032                                                           & 0.080                                                                            & -0.252                                                                        & \textbf{0.418}                                                                   & -0.043                                                                        & 0.051                                                          \\
TNSCUI   & IC-BJC\cite{imagenetc}    & 0.015                                                           & 0.045                                                                            & -0.112                                                                        & 0.188                                                                            & -0.021                                                                        & 0.081                                                          \\
         & Color\cite{2015Structure,2019Fast}   & 0.008                                                           & 0.012                                                                            & -0.015                                                                        & 0.039                                                                            & 0.000                                                                         & 0.018                                                          \\
         & Our & \textbf{0.311}                                                  & \textbf{0.330}                                                                   & \textbf{0.256}                                                                & 0.331                                                                            & \textbf{0.300}                                                                & \textbf{0.355}                                                 \\ \bottomrule
\end{tabular}
\end{table*}

\subsubsection{Implementation Details} \label{subsec:details} 
\textit{\textbf{Network architecture.}} For the KlotskiNet training in DCCL and TNSCUI datasets, we use a pre-trained ResNet-50 \cite{resnet} as the backbone, and replace the top layers with a fully-connected layer and  a binary classification softmax layer.

\noindent
\textit{\textbf{Image split.}}
In image splitting(Line 2 in \cref{algo:KlotskiNetTrain,algo:KlotskiNetInfer}), for an image from DCCL and TNSCUI, we cut it into $5 \times 3$ tiles (each tile has $400 \times 400$ pixels). 

Specifically, in DCCL and TNSCUI, lesions are foreground as well as well annotated. For a tile containing the lesion, it is recognized as a positive tile. 

\noindent
\textit{\textbf{Hyper-parameters.} }
All networks are trained in an end-to-end fashion. Specifically, we utilize stochastic gradient descent with a learning rate 0.001 for DCCL as well as 0.0025 for TNSCUI.
The batch size is set to 32. We use Keras \cite{chollet2015keras} for all our experiments.
Both in Color and our method, $\theta = 12\%$. For IC, the corruption severity is set to 1, which ensures true features are not lost.

\subsection{Overall Performance}\label{subsec:ovelperf}

As shown in \cref{tab:overall}, we observe significant performance drop on our method, which indicates that the biased samples contain strong priors information that the rest samples do not have. Furthermore, this shows that the bias attributes extracted by our method are available in a real prediction model.

\wanglj{Besides, we compare our framework with two baseline methods. Although the drop of negative precision in \tns is worse than IC-BJ, in most cases, our method outperforms. 
Interestingly, we also observe the two baseline methods perform a negatively drop of performance on some scenarios, which indicates these methods are not sufficiently robust to bias validation.}
For color, we investigate the reason for this phenomenon is some of the three channels are not bias attribute at all, top-$\theta$ percentile samples are just outliers and hard to fit. As a result, we suggest more subtle adjustments can be applied on colors as a following studies of \cite{2015Structure,2019Fast}.
\wanglj{For IC, we investigate the corruptions are not necessarily well debiased, but change the distribution of bias features.
For example, low brightness of an image is a negative bias,  darkening of the image makes the model more biased to be negative, leading to a recall drop while a precision raise. As a result, we suggest apply our frameworks to identify the bias
for targeted corruption.
}

\subsection{Performance of Individual Attribute} \label{subsec:ablation}
In this subsection, we show the statistics on performance of individual attributes. Besides to use \klo only, we set Principal Component Analysis(PCA) as a baseline of \bda.  In the following discussions, we use K to represent \klo, K+P to represent \klo with PCA.  We collect the performance drop on each attribute and the statistics are shown in \cref{fig:BoxDCCL,fig:BoxTNSCUI} for DCCL and TNSCUI, respectively. CI-MNIST's is in Appendix.

As shown in \cref{fig:BoxDCCL,fig:BoxTNSCUI}, for KlotskiNet(K) only, we select biased samples based on each component of hidden layer outputs as a direction, the results become unpredictable.
This is due to only valid attributes can be used to select biased samples, while it is difficult to ensure that each output component is valid.  As a result, the selection is randomness, some outliers hard to classify are picked, and finally the result has a reverse performance drop. 

Moreover, PCA(K+P) works better than KlotskiNet(K) only, because it helps to transform the space with respect to the direction by optimizing the within-label variance. That is to say, PCA reduces the randomness in selecting the samples. However, as PCA does not consider the distribution of the other label, the performance drop is not as good as ours, especially, the variance of K+P is larger than ours.

\subsection{Case Studies} \label{subsec:casestudy}

In this subsection, we apply the class activation mapping(CAM)~\cite{gradcam} to visualize the key features recognized by models. Besides \klo and \mil, we train a new \mil model with a debiased training set with respected to a given attribute.  The method to generate the debiased training set is to apply the \textit{project, select and put} operations in \cref{subsec:metric} on the training set to identify biased training samples with $\theta$ as 5\%. We call the model as Debaised. In addition, note that not every bias attribute has a semantics meaning as shown in the cases, because attributes are determined as directions in the high-dimensional space. 

\begin{figure}[t]
   \centering
   \includegraphics[width=0.75\linewidth]{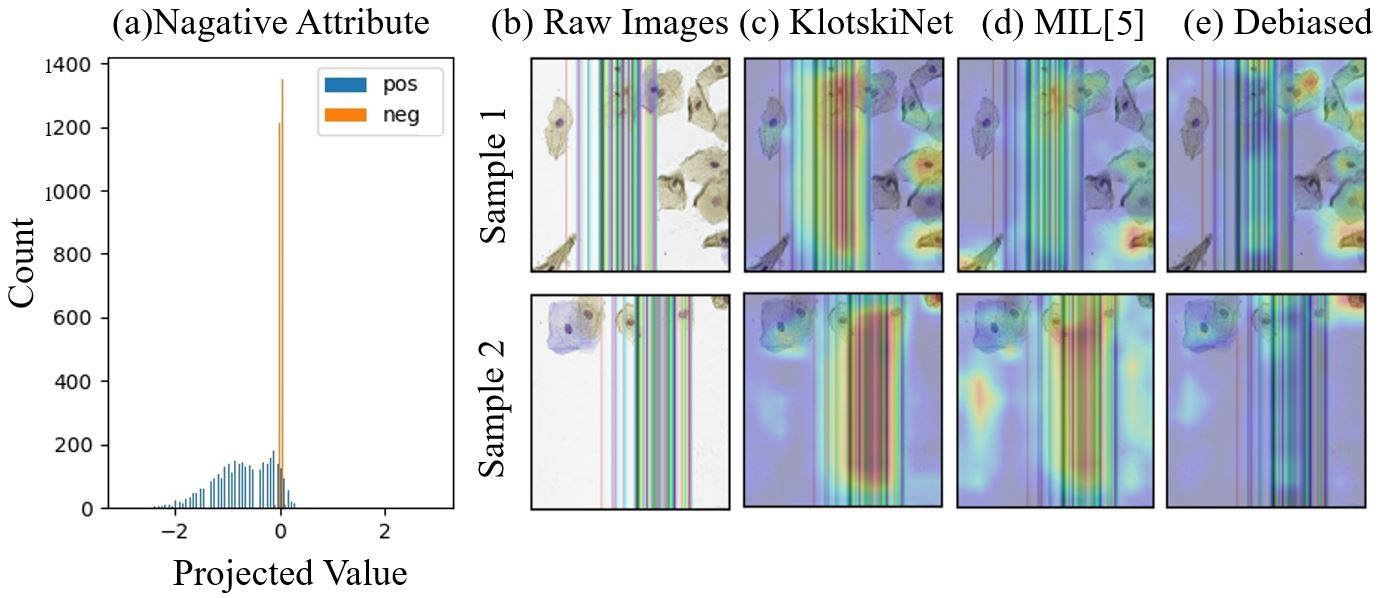}
    \caption{Case Study on DCCL: (a) histogram of projected values; (b) raw images; (c-e) heatmaps from corresponding models}
   \label{fig:caseDCCL}
\end{figure}

\begin{figure}[t]
   \centering
   \includegraphics[width=0.75\linewidth]{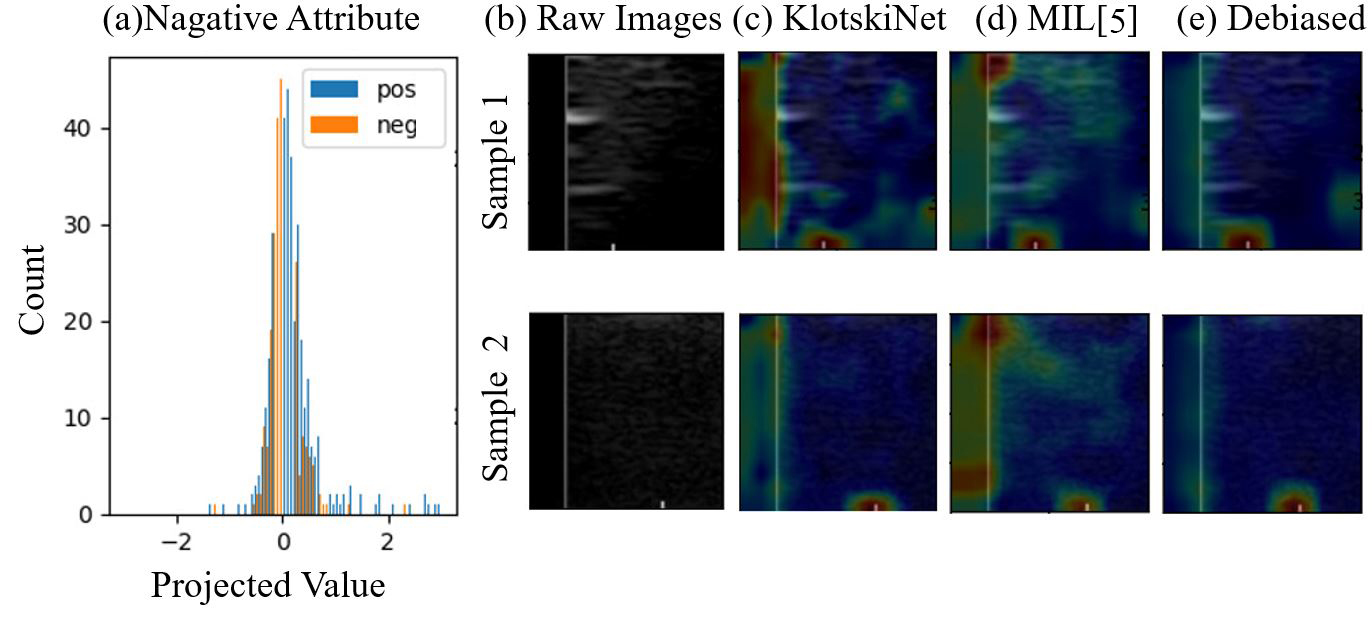}
   \caption{Case Study on \tns: (a) histogram of projected values; (b) raw images; (c-e) heatmaps from corresponding models}
   \label{fig:caseTNSCUI}
\end{figure}

The cases are shown in \cref{fig:caseDCCL,fig:caseTNSCUI} for DCCL and \tns respectively.  In each figure, besides the two raw test images as well as the corresponding images with the heatmap generated from models: \klo, \mil and Debiased (i.e. (b)-(e)), we also demonstrate the distributions of the projected values on the given attribute from positive test sets and negative test sets (i.e (a)).   

Particularly, as shown in \cref{fig:caseDCCL} (b-e), both KlotskiNet and \mil focus on colored stripes while the important features of Debiased are most on cells. %
Similar phenomenon is shown in \cref{fig:caseTNSCUI} (b-e),  both KlotskiNet and \mil focus on the white vertical line in the edge of images, while Debiased almost ignores it. However, there is another bias existing in the two images we selected, which is the white dot at the bottom of the images.  As the bias attribute we selected is to describe the vertical line, and constrained as \cref{con:constrains}, the white dot becomes of the important feature in the Debaised model only ignored the vertical line as shown in \cref{fig:caseTNSCUI} (e).

\begin{figure*}[t]
   \centering
   \includegraphics[width=0.75\linewidth]{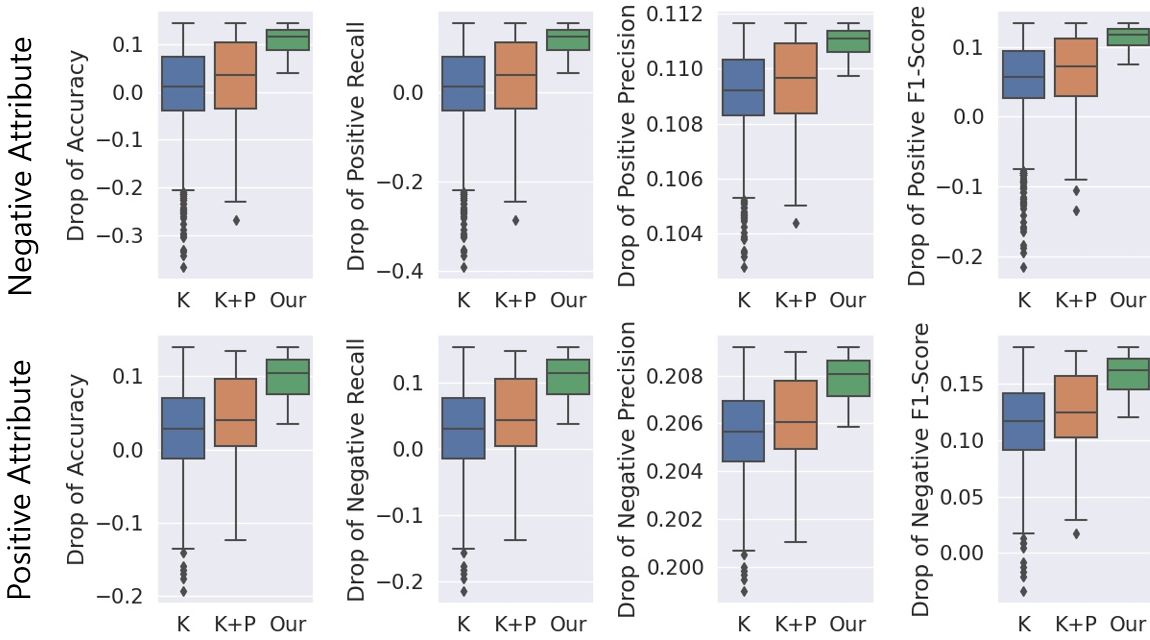}
   \caption{Statistics on Performance of Individual Attributes with DCCL
   }
   \label{fig:BoxDCCL}
\end{figure*}

\begin{figure*}[t]
   \centering
   \includegraphics[width=0.75\linewidth]{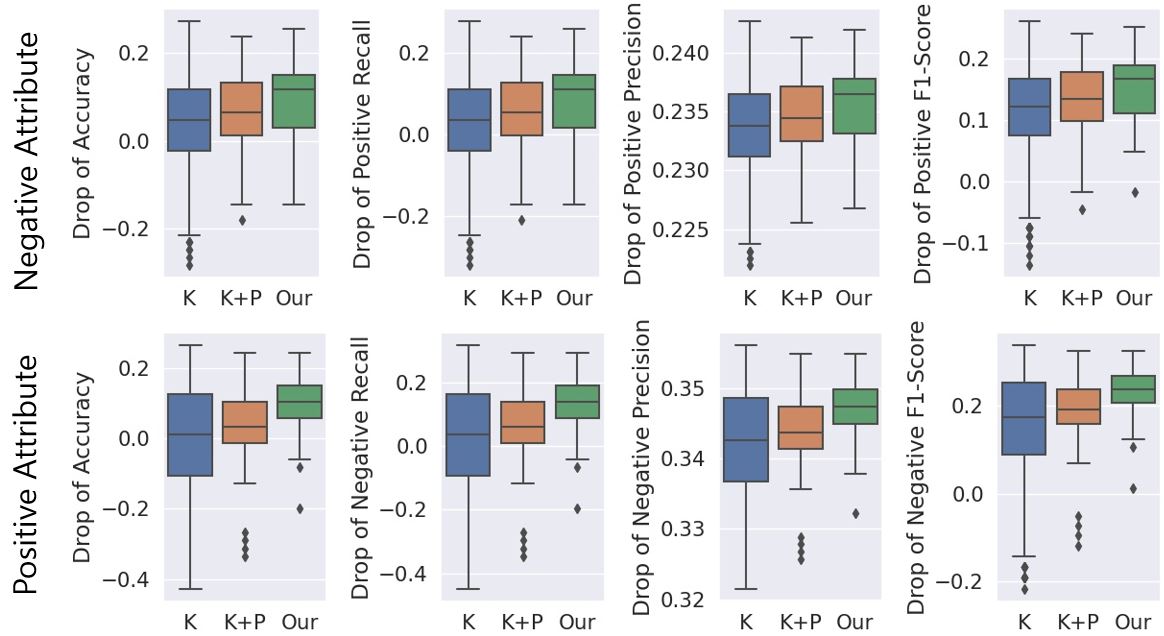}
   \caption{Statistics on  Performance of Individual Attributes with TNSCUI %
   }
   \label{fig:BoxTNSCUI}
\end{figure*}

\section{Conclusion}
This study tackles the issues on automatic bias identification on medical image analysis. We first define the data intrinsic bias attribute and then propose a novel bias identification framework. The framework contains two major components, where KlostkiNet is to build the mapping which makes backgrounds to distinguish positive and negative samples and \bda provides a theoretical solution on determining bias attributes. Extensive experiments show the framework is effective because performance drops without the selected biased samples in a practice model.

%% file: Sections/06_appendix.tex
\section{Appendix}
\subsection{Datasets}
In this section, we provide details of the datasets and settings used in \cref{sec:exp}. 

\begin{table}[]
\centering
\caption{Datasets Overview}\label{tab:datasets}
\begin{tabular}{@{}lrrrrr@{}}
\toprule
           & Train\# & Val\#   & Test\#  & Total\# & Accuracy \\ \midrule
DCCL       & 5,919  & 1,905  & 2,979  & 10,803 & 0.869           \\
TNSCUI & 2,186  & 729   & 729   & 3,644  & 0.750           \\
CI-MNIST  & 50,000 & 10,000 & 10,000 & 70,000 & 0.897           \\ \bottomrule
\end{tabular}
\end{table}

\textbf{DCCL:}
Deep Cervical Cytological Lesions (DCCL) is the largest cervical cytology dataset \cite{dataDCCL}.
We use the provided train, validation, and test splits (5919, 1905, and 2979 images respectively).
The dataset is annotated by six pathologists with eight years of experience on the average on cell level labels: 27,972 lesion cells labeled based on The 2014 Bethesda System.

\textbf{TNSCUI:}
Thyroid Nodule Segmentation and Classification in Ultrasound Images 2020 (referred to as TNSCUI) is the biggest public dataset of thyroid nodule ultrasound images \cite{dataTNSCUI}. 
Each ultrasound image is provided with a bounding box of the nodule and its annotated class (benign or malignant). 
We select a subset of images for a train, validation and test
split of 2186, 729, and 729 images respectively. 

\textbf{CI-MNIST:}
CI-MNIST is a recent published variant of MNIST dataset with introduced different types of biased features \cite{reddy2021benchmarking}.
For an input image , the label  indicates positive or negative, respectively, given that  is even or odd. The train, validation and test splits has 50000, 10000, and 10000 images respectively.
The pipeline of KlotskiNet in CI-MNIST is different from the above mentioned medical datasets. Any tiles containing a part of digits are removed, no matter in a negative or positive sample. To extract the region containing the digit, we implement a simple segmentation algorithm. 

\subsection{Experiment of CI-MNIST}

For the CI-MNIST task, we use three convolution layers with a kernal size $3\times 3$ and the numbers of kernels are 64, 256 and 2,048 followed by a global average pooling layer as the backbone.
In image splitting, for CI-MNIST, we cut it into $4 \times 4$ tiles (each tile has $8 \times 8$ pixels). We regard the region with the digit as the foreground where pixel values of all channels are not zero.  

\subsection{Hyper-Parameters Analysis} \label{subsec:parameter}
\subsubsection{Dimension of Subspace $k$}
\begin{figure}
\centering  
\subfigure[Negative]{\includegraphics[width=0.45\linewidth]{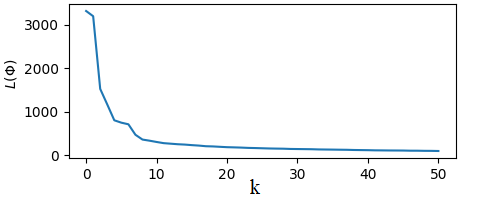}}
\hspace{0.5cm}
\subfigure[Positive]{\includegraphics[width=0.45\linewidth]{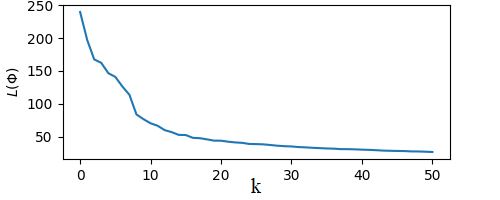}}
\caption{$L(\Phi) - k$ curve of DCCL }\label{fig:lossCurveDCCL}
\end{figure}

\begin{figure}
\centering  
\subfigure[Negative]{\includegraphics[width=0.45\linewidth]{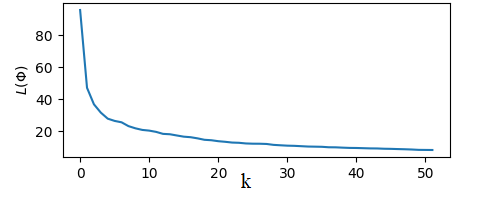}}
\hspace{0.5cm}
\subfigure[Positive]{\includegraphics[width=0.45\linewidth]{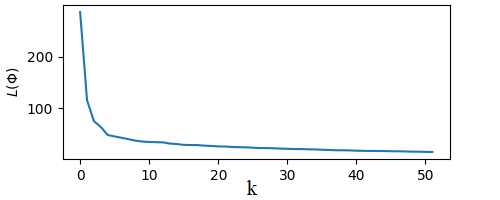}}
\caption{$L(\Phi) - k$ curve of TNSCUI}\label{fig:lossCurveTNSCUI}
\end{figure}

As described in \cref{subsec:bdv}, $\Phi$ can be determined by iterative \cref{con:sol}.
Obviously, the dimension of $\Phi$, is limited by the number of real bias features and the dimension of embedding $\mathcal{U}$ which we set as 2048 in \cref{sec:exp}.
As shown in \cref{fig:lossCurveDCCL,fig:lossCurveTNSCUI}, with the increase of $k$, the value of criterion function $L$ defined in \cref{con:object_func} decreases gradually. We observe that beyond a certain value of $k$, the contribution of the $\phi_k$ is limited and stable, which means it does not contribute on distinguishing positives and negatives. In this study, $k=50$.

In addition, since bias for CI-MNIST datasets is generated of single factors, the maximum of $k$ is 1.

\subsubsection{Threshold on Bias Sample Selection $\theta$}
\begin{figure}
\centering 
\subfigure[DCCL]{\includegraphics[width=0.45\linewidth]{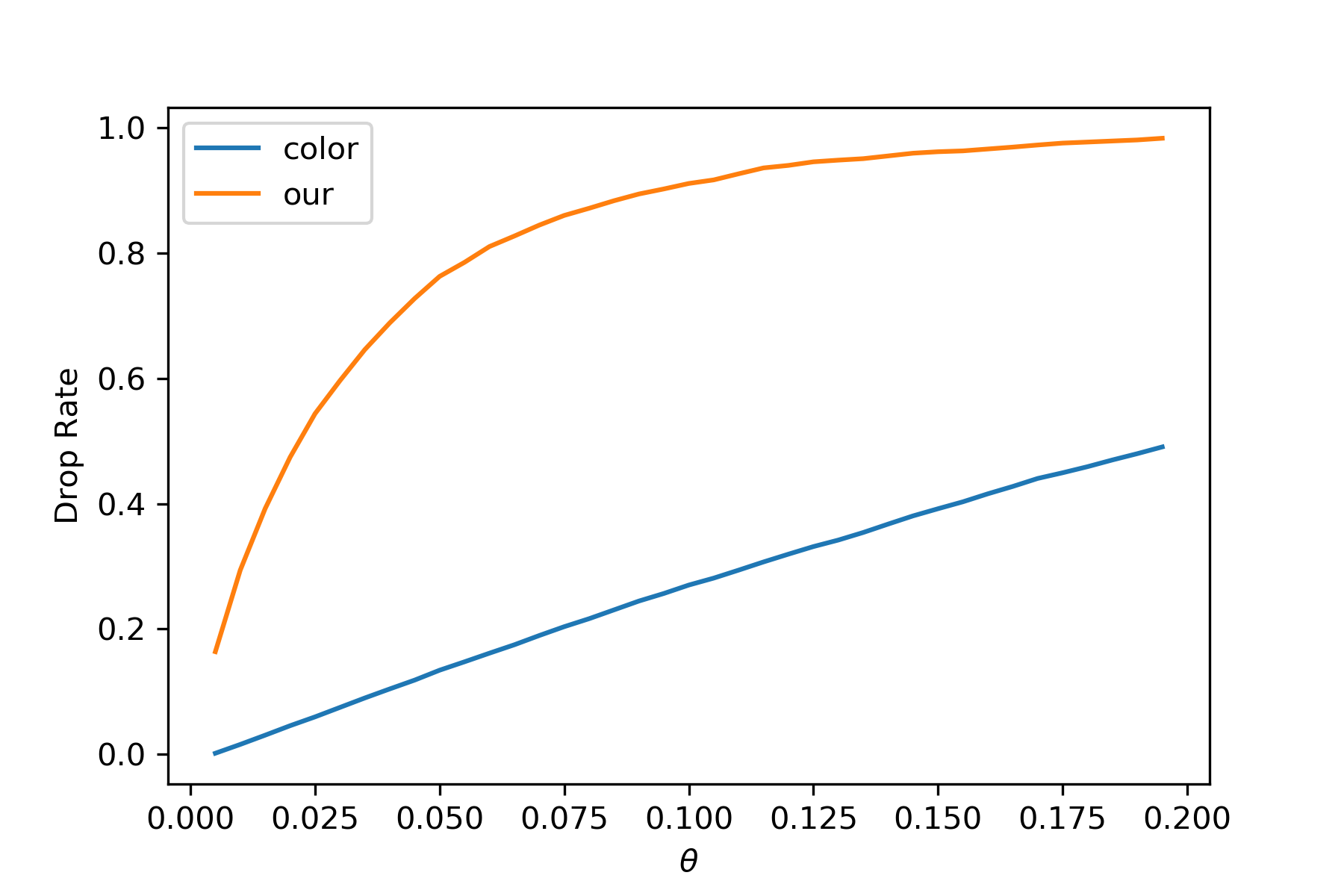}}
\hspace{0.5cm}
\subfigure[TNSCUI]{\includegraphics[width=0.45\linewidth]{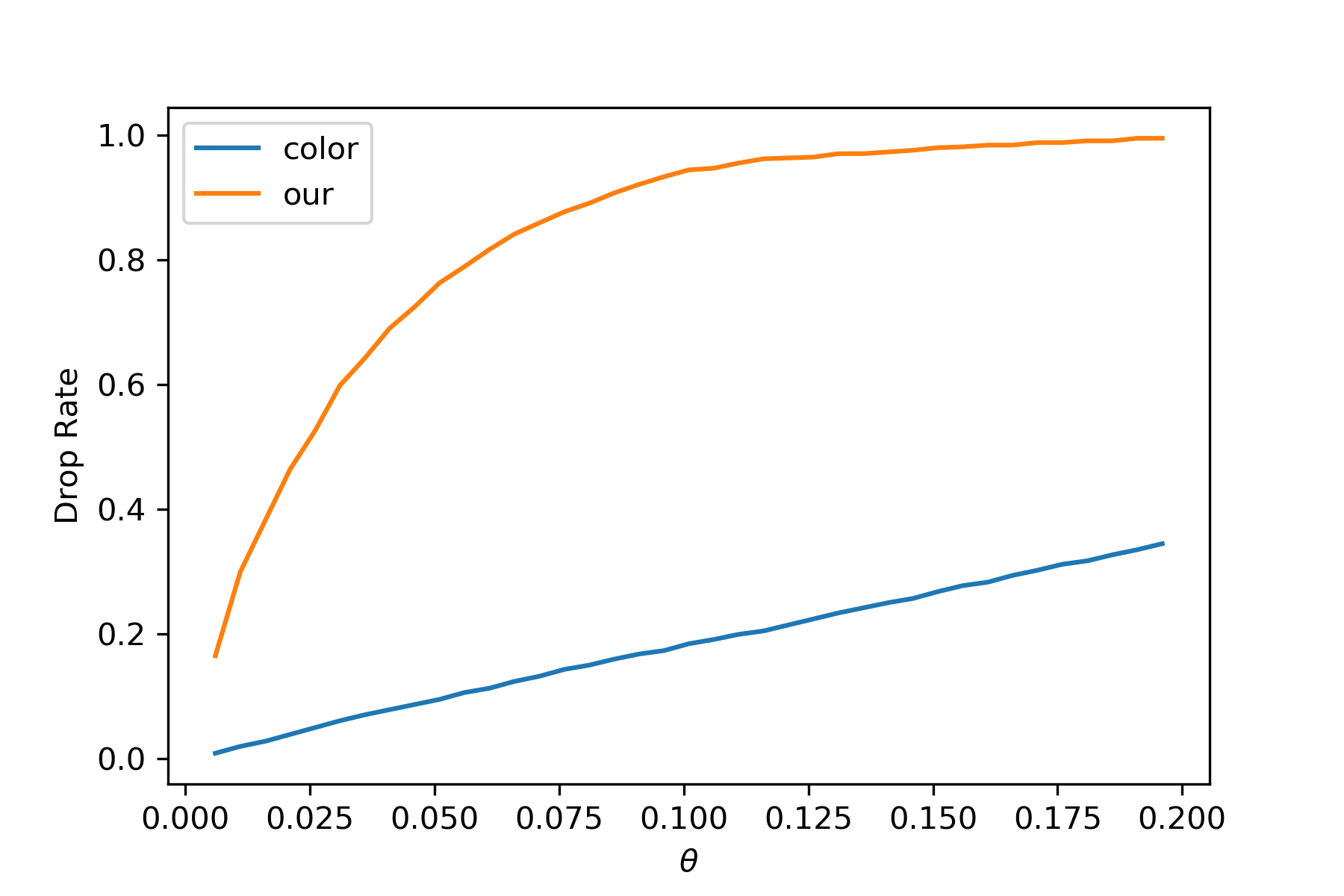}}
\caption{$\theta - drop$ curve of DCCL and TNSCUI }\label{fig:thetaAndDropRate}
\end{figure}

\begin{figure}
\centering  
\subfigure[DCCL]{\includegraphics[width=0.7\linewidth]{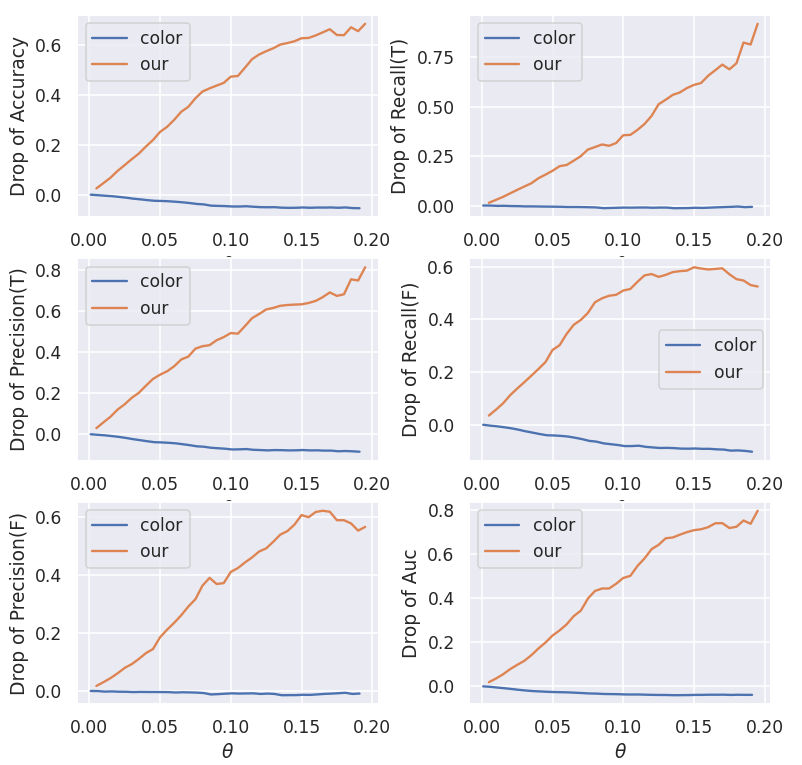}}
\hspace{0.5cm}
\subfigure[TNSCUI]{\includegraphics[width=0.7\linewidth]{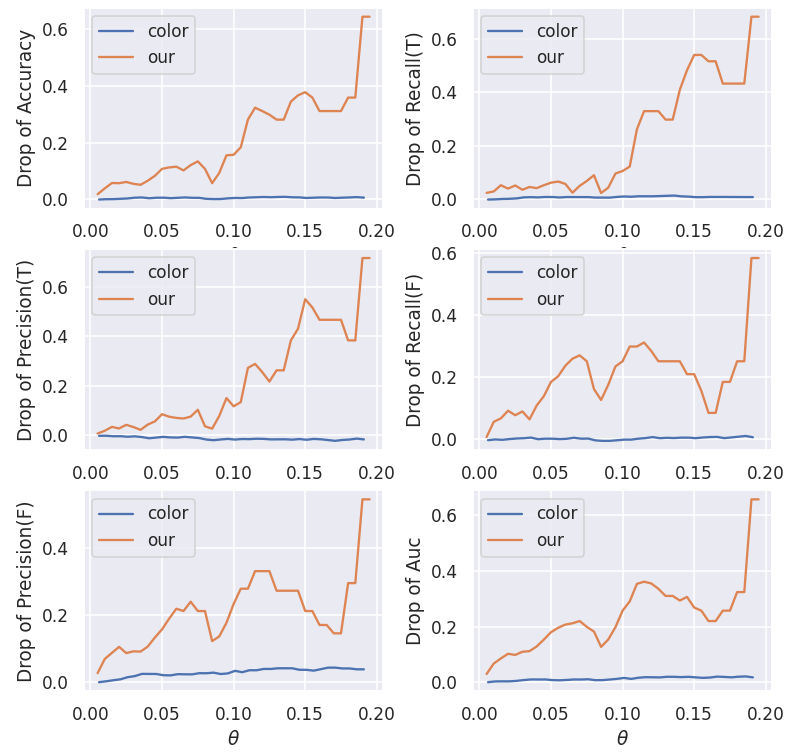}}
\caption{$\theta - performance \quad drop$ curve of TNSCUI}\label{fig:thetaAndDropPreformance}
\end{figure}

As described in \cref{subsec:metric}, $\theta$ is a parameter of data cleansing method, which is used in our method as well as the baseline Color. As shown in \cref{fig:thetaAndDropRate} and \cref{fig:thetaAndDropPreformance}, With the increase of $\theta$, the drop of performance is improved (debiased performance raised) while more data is deleted from the datasets (the available data amount reduced). The trade-off between the debiased performance and the available data amount depends on the actual application.
More $\theta$ extension table of overall performance are shown in the \cref{tab:perext}.
\begin{table}[]
\centering
\caption{Overall Performance Extension Table}\label{tab:perext}
\begin{tabular}{@{}llllllll@{}}
\toprule
       &              & \begin{tabular}[c]{@{}l@{}}Drop of\\      Accuracy\end{tabular} & \begin{tabular}[c]{@{}l@{}}Drop of\\      Positive\\      Precision\end{tabular} & \begin{tabular}[c]{@{}l@{}}Drop of\\      Positive\\      Recall\end{tabular} & \begin{tabular}[c]{@{}l@{}}Drop of\\      Negative\\      Precision\end{tabular} & \begin{tabular}[c]{@{}l@{}}Drop of\\      Negative\\      Recall\end{tabular} & \begin{tabular}[c]{@{}l@{}}Drop of\\      ROC-AUC\end{tabular} \\ \midrule
DCCL   & IC-B         & 0.124                                                           & -0.026                                                                           & 0.213                                                                         & -0.042                                                                           & 0.129                                                                         & 0.043                                                          \\
       & IC-BJ        & 0.174                                                           & 0.201                                                                            & 0.011                                                                         & 0.447                                                                            & 0.105                                                                         & 0.205                                                          \\
       & IC-BJC       & 0.198                                                           & 0.123                                                                            & 0.209                                                                         & 0.167                                                                            & 0.176                                                                         & 0.220                                                          \\
       & Color($\theta=0.02$)   & -0.009                                                          & -0.003                                                                           & -0.014                                                                        & -0.014                                                                           & -0.002                                                                        & -0.013                                                         \\
       & Color($\theta=0.06$)   & -0.029                                                          & -0.008                                                                           & -0.045                                                                        & -0.046                                                                           & -0.005                                                                        & -0.028                                                         \\
       & Color($\theta=0.12$)   & -0.050                                                          & -0.011                                                                           & -0.078                                                                        & -0.087                                                                           & -0.010                                                                        & -0.039                                                         \\
       & Our($\theta=0.02$) & 0.095                                                           & 0.062                                                                            & 0.119                                                                         & 0.112                                                                            & 0.061                                                                         & 0.078                                                          \\
       & Our($\theta=0.06$) & 0.299                                                           & 0.205                                                                            & 0.330                                                                         & 0.345                                                                            & 0.235                                                                         & 0.282                                                          \\
       & Our($\theta=0.12$) & \textbf{0.561}                                                  & \textbf{0.454}                                                                   & \textbf{0.584}                                                                & \textbf{0.573}                                                                   & \textbf{0.480}                                                                & \textbf{0.624}                                                 \\ \midrule
TNSCUI & IC-B         & -0.023                                                          & 0.021                                                                            & -0.150                                                                        & 0.149                                                                            & -0.051                                                                        & -0.018                                                         \\
       & IC-BJ        & 0.032                                                           & 0.080                                                                            & -0.252                                                                        & \textbf{0.418}                                                                   & -0.043                                                                        & 0.051                                                          \\
       & IC-BJC       & 0.015                                                           & 0.045                                                                            & -0.112                                                                        & 0.188                                                                            & -0.021                                                                        & 0.081                                                          \\
       & Color($\theta=0.02$)   & 0.001                                                           & 0.003                                                                            & -0.004                                                                        & 0.008                                                                            & 0.000                                                                         & 0.003                                                          \\
       & Color($\theta=0.06$)   & 0.005                                                           & 0.009                                                                            & -0.009                                                                        & 0.024                                                                            & 0.001                                                                         & 0.008                                                          \\
       & Color($\theta=0.12$)   & 0.008                                                           & 0.012                                                                            & -0.015                                                                        & 0.039                                                                            & 0.000                                                                         & 0.018                                                          \\
       & Our($\theta=0.02$) & 0.056                                                           & 0.041                                                                            & 0.028                                                                         & 0.105                                                                            & 0.035                                                                         & 0.102                                                          \\
       & Our($\theta=0.06$) & 0.115                                                           & 0.058                                                                            & 0.070                                                                         & 0.219                                                                            & 0.064                                                                         & 0.207                                                          \\
       & Our($\theta=0.12$) & \textbf{0.311}                                                  & \textbf{0.330}                                                                   & \textbf{0.256}                                                                & 0.331                                                                            & \textbf{0.300}                                                                & \textbf{0.355}                                                 \\ \bottomrule
\end{tabular}
\end{table}

\subsubsection{Corruption Severity}
\begin{figure}
\centering  
\subfigure[Sample 1 origin]{\includegraphics[width=0.3\linewidth]{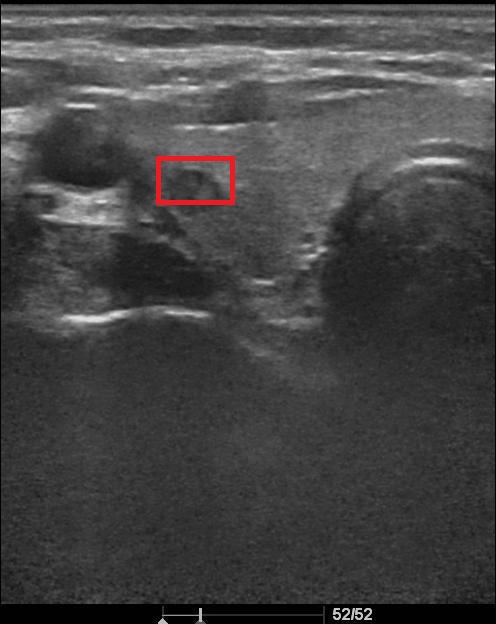}} 
\subfigure[Sample 1 corrupted]{\includegraphics[width=0.3\linewidth]{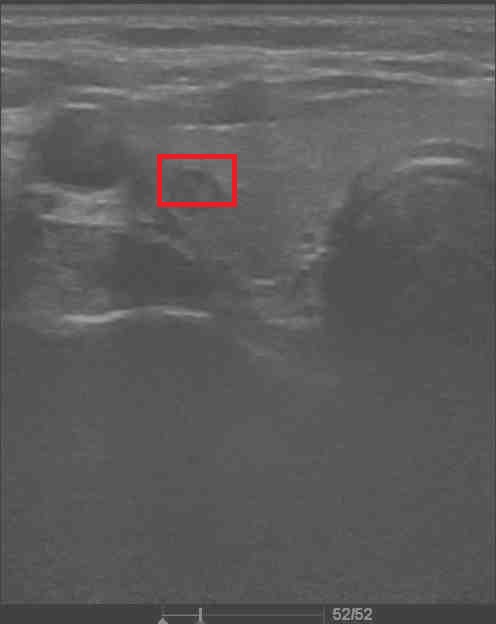}}

\subfigure[Sample 2 origin]{\includegraphics[width=0.3\linewidth]{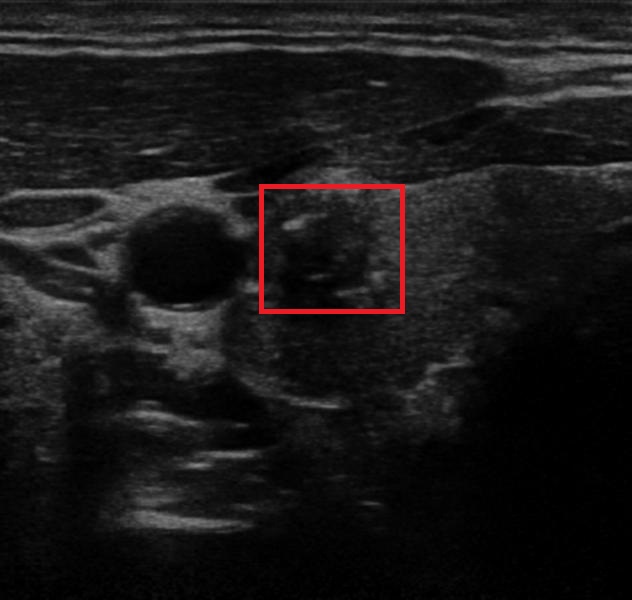}} 
\subfigure[Sample 2 corrupted]{\includegraphics[width=0.3\linewidth]{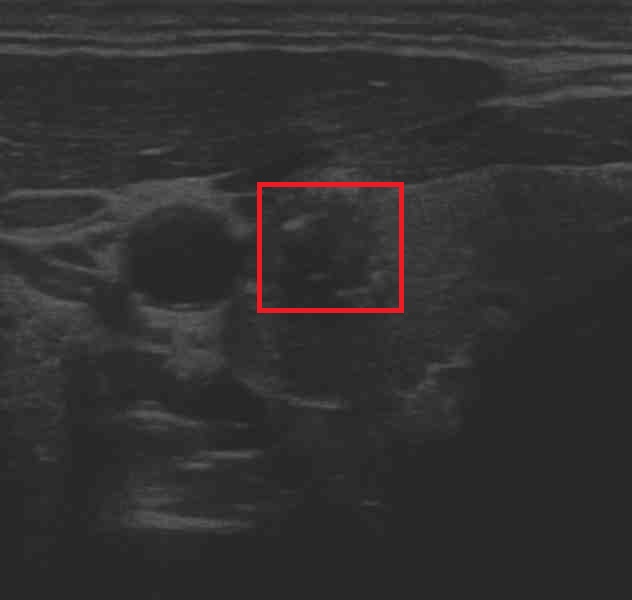}}
\caption{Corruption Severity}\label{fig:corruption}
\end{figure}

As described in the \cite{imagenetc}, each type of corruption has five levels of severity. In this study, corruption severity level = 1.
Corruption severity level $>1$ may cause a reduction of the key evidence related to the target. As shown in \cref{fig:corruption}, a small punctate echogenic foci can be clearly seen in \cref{fig:corruption} (a)\&(c) arrow points area, while the arrow points area in  \cref{fig:corruption} (b)\&(d) seems to be a solid area. This changes could seriously affect radiologist' diagnosis.

\subsection{Individual Performance of CI-MNIST}
\begin{figure*}
\centering  
\subfigure[Negative]{\includegraphics[width=0.9\linewidth]{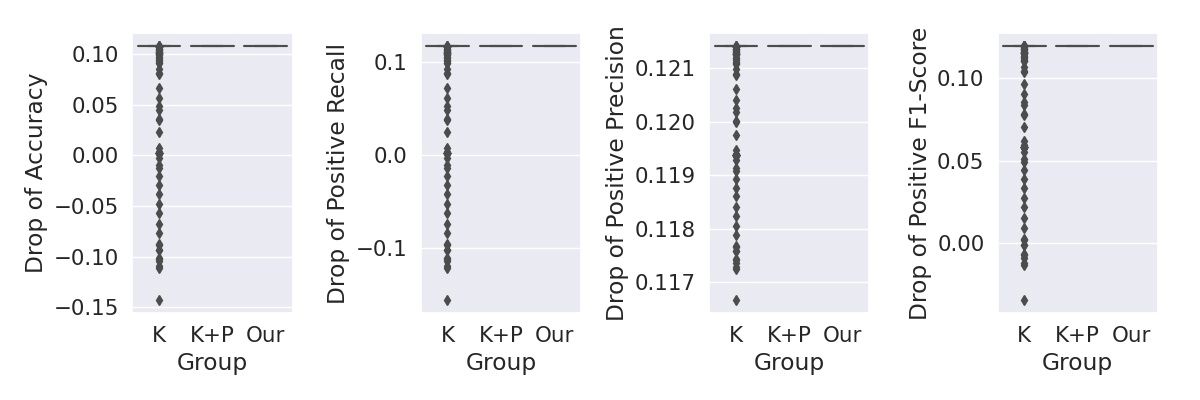}}
\hspace{0.5cm}
\subfigure[Positive]{\includegraphics[width=0.9\linewidth]{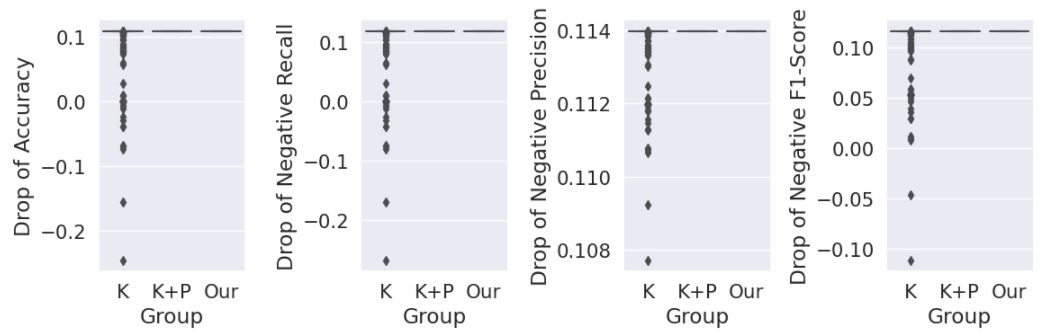}}
\caption{Statistics on  Performance of Individual Attributes with CI-MNIST}\label{fig:boxMNIST}
\end{figure*}

As mentioned in \cref{subsec:parameter},
CI-MNIST datasets is generated of single biased factors, so we only leverage the first component  as the bias attribute from  PCA and \bda, respectively. 
 As indicated in \cref{fig:boxMNIST}, the \bda does not unfold an advantage by comparing with PCA, possibly because of the bias is too simplistic and monolithic in CI-MNIST.

\begin{table}[]
\centering
\caption{Debiased datasets file names}\label{tab:filenames}
\begin{tabular}{@{}lll@{}}
\toprule
Datasets & $\theta$ & List File Name             \\ \midrule
TNSCUI   & 0.02  & debiased\_TNSCUI\_theta002 \\
TNSCUI   & 0.06  & debiased\_TNSCUI\_theta006 \\
TNSCUI   & 0.12  & debiased\_TNSCUI\_theta012 \\
DCCL     & 0.02  & debiased\_DCCL\_theta002   \\
DCCL     & 0.06  & debiased\_DCCL\_theta006   \\
DCCL     & 0.12  & debiased\_DCCL\_theta012   \\ \bottomrule
\end{tabular}
\end{table}